\documentclass[11pt,a4paper]{article}
\usepackage[hyperref]{acl2021}
\usepackage{times}
\usepackage{latexsym}
\usepackage{graphicx}
\usepackage{subfig}
\usepackage{amsfonts,amssymb}
\usepackage{bm}
\usepackage{amsmath}
\usepackage{enumitem}
\usepackage[ruled,linesnumbered,vlined]{algorithm2e}

\usepackage{microtype}
\usepackage{booktabs}
\usepackage{threeparttable}
\usepackage{float}
\usepackage{multirow}
\usepackage[misc]{ifsym}
\usepackage[toc,page]{appendix}
\aclfinalcopy 

\DeclareSymbolFont{extraup}{U}{zavm}{m}{n}
\DeclareMathSymbol{\varheart}{\mathalpha}{extraup}{86}
\DeclareMathSymbol{\vardiamond}{\mathalpha}{extraup}{87}

\title{RevCore: Review-augmented Conversational Recommendation}
\author{Yu Lu$^\dag$, \hspace{0.1cm}
Junwei Bao$^{\ddag(\textrm{\Letter})}\thanks{(\textrm{\Letter}) Corresponding Author}$, \hspace{0.1cm}
Yan Song$^{\dag\sharp}$, \hspace{0.1cm}
Zichen Ma$^\dag$,\\
\bf Shuguang Cui$^{\dag\sharp}$,\hspace{0.1cm}
Youzheng Wu$^{\ddag}$,\hspace{0.1cm}
Xiaodong He$^{\ddag}$ \hspace{0.1cm} \\
  $^\dag$The Chinese University of Hong Kong (Shenzhen)\\
  $^{\ddag}$JD AI Research\hspace{0.8cm} $^{\sharp}$Shenzhen Research Institute of Big Data \\
  \tt{$^{\dag}$\{yulu1,zichenma1\}@link.cuhk.edu.cn}\\
  \tt{$^{\ddag}$\{baojunwei,wuyouzheng1,xiaodong.he\}@jd.com}\\
    \tt{$^{\dag\sharp}$\{songyan,shuguangcui\}@cuhk.edu.cn}\\
  }
\date{}

\makeatletter
\def\thanks#1{\protected@xdef\@thanks{\@thanks
        \protect\footnotetext{#1}}}
\makeatother
\begin{document}
\maketitle



\begin{abstract}

Existing conversational recommendation (CR) systems usually suffer from insufficient item information when conducted on short dialogue history and unfamiliar items.
Incorporating external information (e.g., reviews) is a potential solution to alleviate this problem.
Given that reviews often provide a rich and detailed user experience on different interests, they are potential ideal resources for providing high-quality recommendations within an informative conversation.
In this paper, we design a novel end-to-end framework, namely, \underline{Rev}iew-augmented \underline{Co}nversational \underline{Re}commender (\textbf{RevCore}), where reviews are seamlessly incorporated to enrich item information and assist in generating both coherent and informative responses.
In detail, we extract sentiment-consistent reviews, perform review-enriched and entity-based recommendations for item suggestions, as well as use a review-attentive encoder-decoder for response generation.
Experimental results demonstrate the superiority of our approach in yielding better performance on both recommendation and conversation responding.\footnote{Our code will release in \url{https://github.com/JD-AI-Research-NLP/RevCore}.}
\end{abstract}

\section{Introduction}
With the increasing popularity of intelligent assistants in users' daily lives, how to effectively help users find information or finish specific tasks, such as recommendation and booking, has tremendous commercial potential.
Therefore, conversational recommendation (CR) systems have attracted widespread attention for being a tool providing users potential items of interest through dialogue-based interactions.
%
Though existing studies \citep{sun2018conversational, zhang2018towards, lei2020estimation} proposed to integrate recommender and dialogue components for providing user-specific suggestions through conversations, CR remains challengeable because (i) typical dialogues are short and lack sufficient item information for user preference capturing \citep{chen2019towards, zhou2020improving}, and (ii) difficulties exist in generating informative responses with item-related descriptions \citep{shao2017generating, ghazvininejad2018knowledge, wang2019explainable}.
Thus, recently, external information in the form of structured knowledge graphs (KG) is introduced to enhance item representations by using rich entity information in KG \citep{chen2019towards,zhou2020improving}.
While KG-based methods improve CR to some extent, they are still limited in (i) worse versatility resulted from a high cost of KG construction; and (ii) inadequate integration of knowledge and response generation \citep{lin2020generating}.

\begin{figure}[t]
    \centering
    \includegraphics[width=7.4cm]{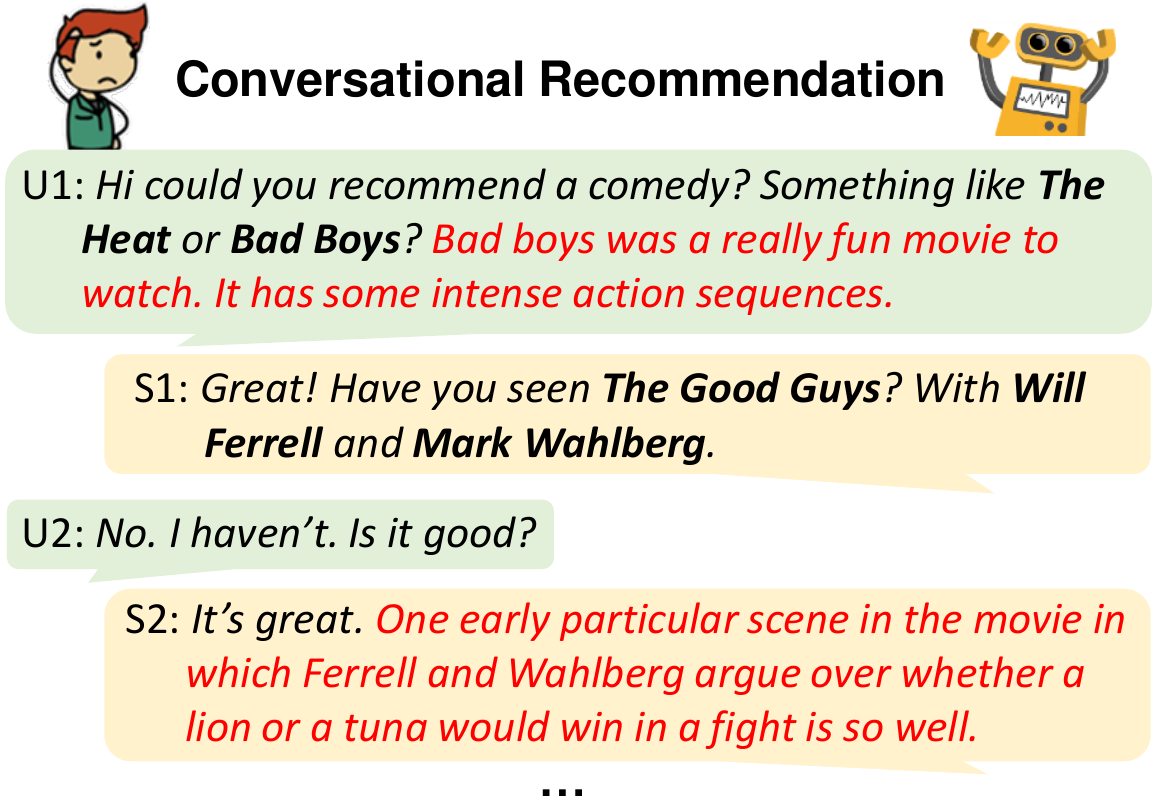}\\
    \caption{An illustrative example of a user-system conversation on movie recommendation. The additional sentiment-matched reviews are in red. Items (movies) and entities (e.g., actors) are in bold.}
    \label{teaser}
\end{figure}

\begin{figure*}[t]
    \centering
    \includegraphics[width=16cm]{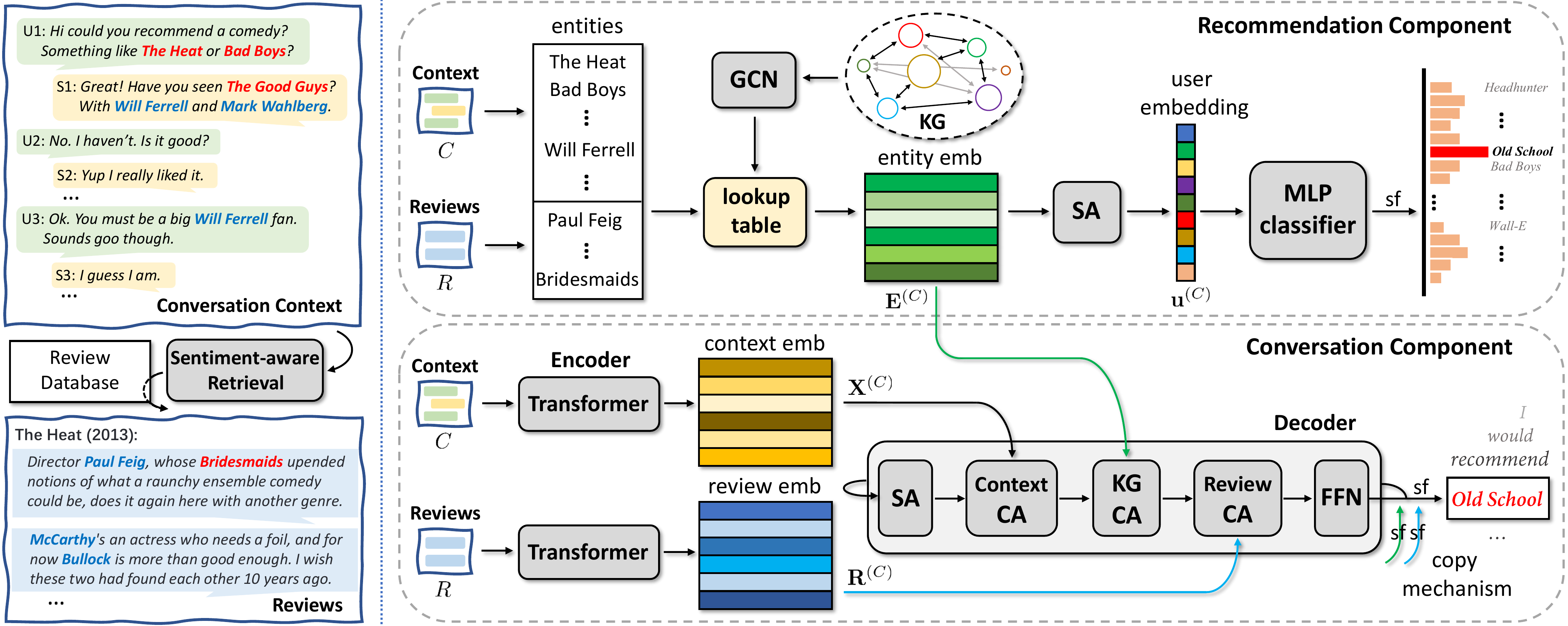}\\
    \caption{The overview of the proposed method in a movie recommendation scenario, where ``emb", ``SA", ``CA", and ``sf" denote embedding, self-attention, cross-attention, and softmax operation, respectively.}
    \label{overview}
\end{figure*}

Given that, nowadays, users are greatly encouraged to share their consumption experience (e.g., restaurant, traveling, movie, etc.), reviews are easily accessed over the internet.
Such reviews often provide rich and detailed user comments on different factors of interest, which are crucial in suggesting recommendations to particular users.
Thus one can treat reviews as promising external sources for higher-quality recommendations in a conversation.
As an example shown in Figure~\ref{teaser}, the CR system may be unfamiliar with the mentioned items from the user, resulting in an uninformative response ``\textit{It's great}.", thus the chat does not help with recommendation owing to lacking necessary knowledge.
In addition, another factor resulting in users' lower acceptance rates to the recommendations is that elaborations on the suggestion are seldom given, which can be alleviated with more explanatory or descriptive utterances after referring to reviews.

Therefore, in better linking external knowledge to recommendation in dialogues,
in this paper, we propose a novel framework, \underline{Rev}iew-augmented \underline{Co}nversational \underline{Re}commender (\textbf{RevCore}), to enhance CR by additional review data.
In doing so, we firstly analyze user's utterances with their sentiment polarities and then
retrieve reviews for the items mentioned by the user with keeping their sentiment matching the utterances (e.g., they should be both positive or negative).
The obtained reviews are thus recommendation-beneficial \citep{he2015trirank, hariri2011context} because they are given by the ones who have seen/used and also show interests (or with no interests) in the mentioned items.
Afterward, we incorporate the selected reviews into dialogue history, from which the CR system can learn user preference from review-enriched item information.
In addition, we also use 
the sentiment-coordinated reviews to enhance the dialogue response generation, where
a review-attentive decoder introduces item information from selected reviews to generate coherent and informative responses.
To the best of our knowledge, it is the first time that the aforementioned CR issues have been addressed through incorporating external reviews.
Experimental results on a widely used benchmark dataset \citep{li2018towards} show that RevCore is superior on both recommendation accuracy and conversation quality.
Further analyses are also performed to confirm the effectiveness of RevCore in an appropriate manner of introducing reviews to CR.

\section{The Proposed Framework}
\label{sec:method}
We present the proposed
\underline{Rev}iew-augmented \underline{Co}nversational \underline{Re}commender (\textbf{RevCore})
with its overview illustrated in Figure~\ref{overview},
where there are three main components, i.e., the review retrieval module, the recommendation component, and the conversation component.
The review retrieval module takes a conversation context $C$ as the input and outputs the selected review set $R$ from the review database $\mathcal{R}_{db}$ which contains all reviews.
The context $C = \{s_t\}_{t=1}^N$ consists of all utterances $s_t$ of the dialogue history given by the user and system in turns, and the review set $R$ includes all review sentence $r \in \mathcal{R}_{db}$ retrieved according to the contexts in previous turns.
With $C$ and $R$ as the input, the recommendation component outputs a set of items from the candidate item set $\mathcal{Z}$ as the recommendation.
The dialogue component also accepts $C$ and $R$ as input, and outputs an utterance $s_{t+1} = \{w_i\}_{i=1}^{M}$ as the response, where $w_i$ is the $i_{\rm th}$ word and $M$ the length of $s_{t+1}$.
The output $s_{t+1}$ is added to the context of the next turn.

We first introduce how to retrieve proper reviews from a database in the following Section~\ref{sec:retrieval}.
Then our solutions to the recommendation and conversation tasks are described in Section~\ref{sec:rec} and Section~\ref{sec:conv}  respectively, along with detailed illustrations of how reviews enhance both two tasks.
Without the loss of generality, our method is introduced in a movie recommendation scenario.

\subsection{Review Retrieval} 
\label{sec:retrieval}
To help dialogue with reviews, given $\mathcal{R}_{db}$, it is of great importance to retrieve proper ones.
The reasons are two folds: (i) non-relevant reviews may result in harmful effects to the user representation; (ii) reviews with inconsistent altitudes inject noise into the conversation, which impedes generating coherent responses.
{Then, a preliminary retrieval is to search in $\mathcal{R}_{db}$ for proper reviews according to the mentioned item in the conversation context $C$.}
{For review filtering, we design a sentiment-aware retrieval module.}
{The sentiment value $v \in [0, 1]$ of each review $r$ can be captured by a transformer-based sentiment predictor:}
\begin{equation}
\setlength{\abovedisplayskip}{6pt}
\setlength{\belowdisplayskip}{6pt}
    v~ = ~{\rm Sentiment}(r), ~r \in \mathcal{R}_{db},
    \label{equ:senti}
\end{equation}
{where ${\rm Sentiment}(\cdot)$ denotes sentiment prediction, and $v$ can be viewed as how well the movie is liked in this review $r$.}
{Similarly, the sentiment of a response to this movie can also be obtained in this way.}
{As a result, reviews that possess similar sentiment polarity $v^*$ with the response are selected.}

{Considering helpful reviews are usually long paragraphs, we only retain part of them, one sentence, for each mentioned movie.}
{Given a context $C$, there exist two manners to select the sentences $r^{(C)}$ from the raw reviews, word-wisely (or phrase-wisely) and sentence-wisely.}
{The first one randomly chooses some words or phrases to form each ``sentence", and one whole sentence is directly selected at random in a second way.}
{Despite the expense of sentence fluency, the first manner enjoys much variability due to the extensive word/phrase combinations.}
The process to obtain $r^{(C)}$ can be formulated as follows:
\begin{equation}
\setlength{\abovedisplayskip}{6pt}
\setlength{\belowdisplayskip}{6pt}
    r^{(C)} = ~{\rm Retrieve}(\mathcal{R}_{db}, V, v^*),
    \label{equ:retr}
\end{equation}
{where ${\rm Retrieve}(\cdot)$ denotes the retrieval operation and $V$ is the set of all $v$.}
{The obtained $r^{(C)}$ is added into the review set $R$.}

{With the retrieved review sentence, one way of incorporation is to briefly insert it right behind the sentence where the item is, as in Fig.~\ref{teaser}.}
{However, it may cause the perturbation to the conversational consistency by interrupting the original dialogue.}
{Thus we seamlessly incorporate the review embedding into the conversation component, which is described in Section.~\ref{sec:conv}.}
{More importantly, the review sentence serves as a brief introduction or explanation to the mentioned movie.}
{It enriches user information for personalized recommendations and introduces external knowledge for more informative recommendation responses.}


\subsection{Review-augmented Recommendation} 
\label{sec:rec}
{The recommender component is constructed based on a KG-based framework \citep{zhou2020improving}, with all entities in the context are extracted to generate the embedding of a user profile. In our method, the retrieved reviews work on enriching entity information so that the user embedding can be augmented to promote recommendation accuracy.}

{Similar to the approach in \citet{zhou2020improving}, a candidate entity embedding dictionary $\mathcal{E}$ is constructed first by using GNN to learn entity representations from KG, e.g., DBpedia \citep{auer2007dbpedia}.}
{Given a context $C$, all entities $E^{(C)}$ in it are extracted. Then the embedding vectors of them are looked up from $\mathcal{E}$ and concatenated into a matrix $\mathbf{E}^{(C)} \in \mathbb{R}^{l^{(C)} \times d}$, where $l^{(C)}$ is the number of entities in the context $C$, and $d$ denotes the embedding dimension.}
{Next, the entity embedding $\mathbf{E}^{(C)}$ is aggregated into a user embedding vector $\mathbf{u}^{(C)}$, through a self-attention layer (SA) as follows:}
\begin{equation}
\setlength{\abovedisplayskip}{6pt}
\setlength{\belowdisplayskip}{6pt}
   \begin{aligned}
      \mathbf{u}^{(C)} &= ~ \mathbf{E}^{(C)} \cdot \bm{\alpha}, \\
      \bm{\alpha}~~ &= ~ {\rm softmax}(\mathbf{b}^\top \cdot {\rm tanh} (\mathbf{W}_\alpha \mathbf{E}^{(C)})),
   \end{aligned}
\end{equation}
{where $\bm{\alpha}$ is the attention weight vector, and $\mathbf{W}_\alpha$ and $\mathbf{b}$ are the parameter matrix and vector for linear projection and bias.}
{Given $\mathbf{u}^{(C)}$, a multi-layer perceptron (MLP) and a softmax operation are adopted to obtain the recommendation prediction $\mathbf{p} \in \mathbb{R}^L$, where $L$ is the number of candidate movies:}
\begin{equation}
\setlength{\abovedisplayskip}{6pt}
\setlength{\belowdisplayskip}{6pt}
\mathbf{p} = {\rm softmax}({\rm MLP} (\mathbf{u}^{(C)})).
\end{equation}
{To learn parameters in the recommender component, a cross-entropy loss $\mathcal{L}_{rec}$ between the prediction $\mathbf{p}$ and the target movie category is computed:}
\begin{equation}
\setlength{\abovedisplayskip}{6pt}
\setlength{\belowdisplayskip}{6pt}
\mathcal{L}_{rec} = -\frac{1}{M}\sum_{i=1}^{M}{\rm log}~p^*_i, 
\label{loss:rec}
\end{equation}
{where $M$ is the number of recommendations and $p^*_i$ is the prediction probability of the target category in the $i_{\rm th}$ recommendation.}
 
{The pipeline described above suffers from the entity sparsity in dialogue history, resulted from the dataset construction process, where annotators are inevitably unfamiliar with some movies.
Retrieved reviews can act to enrich the $E^{(C)}$ by adding more entity words.}
{The process of obtaining review-enriched entities can be formulated as:}
\begin{equation}
\setlength{\abovedisplayskip}{6pt}
\setlength{\belowdisplayskip}{6pt}
   \begin{aligned}
      E_C^{(C)} &= {\rm Extract}(C), \\
      E_R^{(C)} &= {\rm Extract}(R^{(C)}), \\
      E^{(C)} &= \{ E_C^{(C)}, E_R^{(C)}\},
   \end{aligned}
\end{equation}
{where ${\rm extract}(\cdot)$ defines the entity extraction operation, and $E_c^{(C)}$, $E_r^{(C)}$ denotes entities extracted from the context and retrieved review, respectively.}
{Based on the review-enriched entities, the user embedding is expected to be better represented to produce a more precise recommendation.}

\subsection{Review-augmented Response Generation}
\label{sec:conv}
{Reviews can also augment the response generation in the conversation component.}
{We build an encoder-decoder framework to handle the generation task.}
{Retrieved reviews and context are encoded separately first, for the purpose of maintaining the dialog consistency.}
{In the decoding stage, the review embedding is fused via an attention layer to generate informative responses.}
{Considering that a good modeling of the input plays an important role to achieve an outstanding model performance \cite{mikolov2013efficient,ijcai2018-607,peters-etal-2018-deep,devlin2019bert,song2021zen} and transformer-based approaches have achieved state-of-the-art in many NLP tasks \cite{vaswani2017attention,chen2019towards,zhou2020improving,chen-etal-2020-joint,joshi-etal-2020-spanbert,wang-etal-2020-learning-decouple,tian-etal-2020-supertagging}, we adopt two transformers as the encoders for context and reviews.}
{Given a context $C$ and the retrieved reviews $R$, the context embedding $\mathbf{X}^{(C)}$ and review embedding $\mathbf{R}^{(C)}$ are first obtained:}
\begin{equation}
\setlength{\abovedisplayskip}{8pt}
\setlength{\belowdisplayskip}{8pt}
   \begin{aligned}
      \mathbf{X}^{(C)} &= {\rm Transformer}_{\bm \theta_X}(C), \\
      \mathbf{R}^{(C)} &= {\rm Transformer}_{\bm \theta_R}(R), 
   \end{aligned}
\end{equation}
{where ${\bm \theta_X}, {\bm \theta_R}$ are parameters in these two transformers.}
{The decoding stage takes them and the entity embedding $\mathbf{E}^{(C)}$ as inputs of attention layers.}
{These attention layers aim to fuse the external information from KG and reviews $R$ into the context information, inspired by the work of \newcite{zhou2020improving}.}
{Given the decoding output of last time unit $\mathbf{Y}^{i-1}$, the current one $\mathbf{Y}^{i}$ is generated by:}
\begin{equation}
\setlength{\abovedisplayskip}{8pt}
\setlength{\belowdisplayskip}{8pt}
   \begin{aligned}
      \mathbf{A}_0^i &= {\rm MHA}(\mathbf{Y}^{i-1}, \mathbf{Y}^{i-1}, \mathbf{Y}^{i-1}), \\
      \mathbf{A}_1^i &= {\rm MHA}(\mathbf{A}_0^i, \mathbf{X}^{(C)}, \mathbf{X}^{(C)}), \\
      \mathbf{A}_2^i &= {\rm MHA}(\mathbf{A}_1^i, \mathbf{E}^{(C)}, \mathbf{E}^{(C)}), \\
      \mathbf{A}_3^i &= {\rm MHA}(\mathbf{A}_2^i, \mathbf{R}^{(C)}, \mathbf{R}^{(C)}), \\
      \mathbf{Y}^i &= {\rm FFN}(\mathbf{A}_3^i),
   \end{aligned}
   \label{equ:decoder}
\end{equation}
{where ${\rm MHA}(\mathbf{Q}, \mathbf{K}, \mathbf{V})$ represents the multi-head attention function \citep{vaswani2017attention}, which takes a query, key, and value as input:}
\begin{equation}
\setlength{\abovedisplayskip}{8pt}
\setlength{\belowdisplayskip}{8pt}
   \begin{aligned}
      &{\rm MHA}(\mathbf{Q}, \mathbf{K}, \mathbf{V}) = [{\rm h}_1, \cdots, {\rm h}_h ] \cdot \mathbf{W}^o, \\
      &{\rm h}_i = {\rm Attention}(\mathbf{Q}\mathbf{W}_i^{(q)}, \mathbf{K}\mathbf{W}_i^{(k)}, \mathbf{V}\mathbf{W}_i^{(v)}),
   \end{aligned}
\end{equation}
{where $[\cdot]$ represents the concatenation operation, $h$ is the number of heads, and $\mathbf{W}_i$ is the parameter matrix to learn.}
{${\rm FFN}(\cdot)$ in Equation~\ref{equ:decoder} defines a fully-connected feed-forward network, which comprises of two linear layers with one ReLU activation layer in between:}
\begin{equation}
\setlength{\abovedisplayskip}{8pt}
\setlength{\belowdisplayskip}{8pt}
   {\rm FFN}({\bm x}) = {\rm ReLU}({\bm x}\mathbf{W}_1 + \mathbf{b}_1) \mathbf{W}_2 + \mathbf{b}_2. 
\end{equation}
{As presented above, information is injected progressively into the decoding stage, from the original context at first, then related entity information in KG, and finally reviews, which contain detailed item-related information.} 

{To complete the generation, the decoder output $\mathbf{Y}^i$ is processed through a softmax operation to predict the token distribution.}
{Apart from the conversational consistency required in the chit-chat task, the CR system also expects recommendation-related responses, which usually contain relevant entities and descriptive keywords.}
{So a copy mechanism is further adapted to introduce vocabulary bias and thus increase the informativeness in the generation.}
{Given the previous generated sub-sequence $\{y_{i-1}\} = y_1, y_2, \cdots, y_{i-1}$, the generation probability $y_i$ of the next token can be computed as:}
\begin{equation}
\setlength{\abovedisplayskip}{8pt}
\setlength{\belowdisplayskip}{8pt}
 \begin{aligned}
   {\rm Pr}(y_i | \{y_{i-1}\}) = & \  {\rm Pr}_1(y_i | \mathbf{Y}_{i}) + {\rm Pr}_2(y_i | \mathbf{Y}_{i}, G) + \\ & \  {\rm Pr}_3(y_i | \mathbf{Y}_{i}, R), 
 \end{aligned}
\end{equation}
{where ${\rm Pr}_1(\cdot)$ is a generation probability function over the vocabulary, with $\mathbf{Y}^i$ as the input. $G$ and $R$ represents the knowledge graph and reviews we use. ${\rm Pr}_2(\cdot)$, ${\rm Pr}_3(\cdot)$ are copy probability functions from KG entities and reviews, respectively, implemented by a standard copy mechanism \citep{gulcehre2016pointing} (computing the distributions over the KG words or review words).}
{Both probability functions are implemented with a softmax operation. To learn the response generation in the dialogue component, we set a cross-entropy loss:}
\begin{equation}
\setlength{\abovedisplayskip}{8pt}
\setlength{\belowdisplayskip}{8pt}
\hspace{-2pt}
   \mathcal{L}_{gen} = -\frac{1}{N}\sum_{t=1}^{N}{\rm log}\big ({\rm Pr}(s_t|s_1,
   {\small \cdots},s_{t-1})\big ), 
   \label{loss:gen}
\end{equation}
{where $N$ is the number of turns, $s_t$ represents the $t_{\rm th}$ utterance in the conversation.}

{To train the whole model, it includes three steps:
(i) pre-training the sentiment predictor in the review retrieval module; (ii) training the recommender component by minimizing $\mathcal{L}_{rec}$; (iii) training the dialogue component by minimizing $\mathcal{L}_{gen}$.}

\section{Experiment Settings}

\subsection{Dataset} 
REDIAL \citep{li2018towards} is a widely-used dataset of real-world conversations around the theme of providing movie recommendations generated by the human in seeker-recommender pairs.
REDIAL contains 10,021 conversations related to 64,362 movies, split into training, validation, and test sets using a ratio of 8:1:1\footnote{More statistics presents in appendix~\ref{appendixA}.}.
To construct a review database, we crawled 30 reviews for each movie from IMDb\footnote{\url{https://www.imdb.com/}} website, which is one of the most popular and authoritative movie databases. 
Each review can be queried according to the corresponding movie along with its rating and helpful score provided by IMDb. 
{In practice, we select the 30 reviews with the highest helpful scores for each movie to guarantee the high quality of collected reviews.}
{Other manners of selecting the 30 reviews are described and compared in the second part of Section~\ref{diss}.}

\subsection{Implementation Details}
{The maximum lengths of context and response are set to 256 and 30, respectively. Transformers for review encoding in dialogue generation and sentiment prediction use the same hyper-parameters with the context encoder. For sentiment polarity in the reviews, we threshold on the star-rating to getting sentiment polarity with the threshold set to 5. In the dialogue context, the sentiment polarity is obtained according to users' attitude to the mentioned entity in utterances, which is provided by the REDIAL dataset.  Other settings are kept consistent with \newcite{zhou2020improving} for fair comparison\footnote{More details of hyper-parameters and training strategies are described in Appendix~\ref{appendixB}; the size of different models and their inference speed are reported in Appendix~\ref{appendixC}.}.} 
{Besides, the ``review sentence'' is selected according to the sentiment value and in a sentence-wise manner, and the token number of incorporated review sentences is set to 20, considering the balance between the original source and external source. We add the retrieved review sentences after the mentioned items in the dialogue component training to guide it to generate review-aware responses.}
{The sentiment predictor for reviews is trained on the collected reviews. The sentiment predictor for dialog context is trained on the IMDb Movie Reviews Dataset \citep{maas-etal-2011-learning} and then finetuned on the REDIAL dataset.}

\subsection{Baselines}
{Evaluated on the REDIAL dataset, we compare our approach with a variety of competitive baselines from previous studies listed as follows:}
\setlength{\parskip}{0em}
\begin{itemize}[leftmargin=*]
\setlength{\itemsep}{0pt}
\setlength{\parsep}{0pt}
\setlength{\parskip}{1mm}
    \item\textbf{Trans}{\citep{vaswani2017attention} applies a encoder-decoder framework based on transformer for generation, and applies a transformer encoder to encode context information for recommendation.}
    \item \textbf{Redial} {\citep{li2018towards} builds a conversation component based on a hierarchical encoder-decoder architecture, and its recommender component is implemented by an auto-encoder extended with a RNN-based sentiment analysis module.} 
    \item \textbf{KBRD} \citep{chen2019towards} {adopts DBpedia-enhanced contextual items or entities to construct user profile for recommendation. The KG-enhanced user profile also serves as word bias for the transformer-based generation module.}
    \item \textbf{KGSF} {\citep{zhou2020improving} uses MIM \citep{viola1997alignment} to align the semantic spaces of two KGs. The user embedding is obtained from the aligned representations of words and items for recommendation. The generation module follows a transformer encoder and a fused KG enhanced decoder.}

\end{itemize}

\subsection{Evaluation Metrics} 
{Our method is evaluated on both the recommendation and conversation tasks.}
{The evaluation metric for recommendation is Recall@$k$ (R@$k$, $k=$ 1, 10, 50), which indicates whether the predicted top-$k$ items contain the ground truth recommendation provided by human recommenders.}
{Conversation evaluation comprises automatic and human evaluation.}
{The metrics for automatic evaluation are perplexity (PPL) \citep{jelinek1977perplexity}  and distinct $n$-gram (Dist-$n$, $n=$ 2, 3, 4)~\citep{li-etal-2016-diversity}.}
{Perplexity is a measurement for the fluency of natural language, where lower perplexity refers to higher fluency.}
{Distinct $n$-gram is a measurement for the diversity of generated utterances.}
{Specifically, we use distinct 3-gram and 4-gram at the sentence level to evaluate the diversity.}
{The main purpose of our dialog component is a successful recommendation rather than imitating the ground truth responses.
Therefore, we provide annotators to manually evaluate the results instead of using BLEU scores.}
{The annotators evaluate the quality of generated dialogue responses from 3 aspects, i.e., coherence, fluency, and informativeness, with each score ranging from 0 to 1.}

\begin{table}[t]
    \centering
    \setlength{\tabcolsep}{3mm}
    \setlength{\abovecaptionskip}{2mm}
    \begin{tabular}{lccc}
    \toprule
    \small \textbf{Models} & \small \textbf{R@1} & \small \textbf{R@10} & \small \textbf{R@50} \\
    \midrule
    \small Redial & 2.4 & 14.0 & 32.0 \\
    \small KBRD   & 3.1 & 15.0 & 33.6 \\
    \small KGSF   & 3.9 & 18.3 & 37.8  \\ 
    \hline

    \small RevCore ($-$KG) & 4.2 & 22.7 & 43.3 \\
     \small \textbf{RevCore ($+$KG)} &\bf 6.1 &\bf 23.6 & \bf 45.4 \\ 
    \bottomrule
    \end{tabular}
    \caption{ Results on the recommendation task. Best results are in bold. }
    \label{tab:recs}
\end{table}

\section{Results and Analysis}

\subsection{Evaluation on Recommendation Task}
\label{eval:rec}
{For the recommendation task, we adopt Recall@$k$ (R@1, R@10, R@50) for evaluation.}
{As the results summarized in Table~\ref{tab:recs}, our approach outperforms all competitive baselines and achieves 5.9\% R@1, 24.0\% R@10, and 41.3\% R@50, which is the state-of-the-art performance on the REDIAL dataset.\footnote{We report the performance of different models on the validation sets in Appendix~\ref{appendixD} and the mean and standard deviation of the test set results in Appendix~\ref{appendixE}.}}
{Compared with KGSF, RevCore ($+$KG) achieves significant improvements, with R@1 score improved about 156\% (absolutely 2.2), R@50 score improved about 129\% (absolutely 4.5), and R@50 score improved about 120\% (absolutely 7.6).}

{We also evaluate the performance of RevCore ($-$KG), which means the construction of $\mathcal{E}$ removes relation between entities.} 
{Instead, an embedding matrix is randomly initialized and learned to represent each entity, without using the GNN-based embedding.}
{In this version, the external knowledge source we introduce is reduced to review only. As the result in the last two rows of Table~\ref{tab:recs}, RevCore  ($-$KG) can achieve competitive results with RevCore ($+$KG), and outperform KGSF that uses two KGs.}
{According to our observation, although the learning of entity representation is made harder without structured knowledge graphs, the enrichment of dialogue history by reviews makes up the embedding learning.}
{It demonstrates that incorporating reviews is a meaningful method to improve the recommendation in the conversation. We hope this result inspire further research.}


\subsection{Evaluation on Conversation Task}
\label{eval:con}
\paragraph{Automatic Evaluation}
{The results of automatic evaluation on the REDIAL dataset summarize in Table~\ref{tab:conv}.}
{The proposed RevCore outperforms all competitive baselines and achieves significant improvements over most of the automatic metrics.}
{Compared with KGSF, all of the Dist-$n$ scores are significantly lifted, namely, by +0.14 for Dist-2, +0.11 for Dist-3, and +0.08 for Dist-4, which demonstrates our method is effective to generate diverse utterances.}
{Besides, RevCore ($+$KG) achieves a comparable PPL score with KGSF.}
{It validates our claim that the review incorporation in our method does not cause a decline in generation fluency.}
{The lower PPL score of RevCore ($+$KG) possibly relates to the high fluency contained in incorporated reviews that carefully induct by website users.}
{For the version of RevCore ($-$KG), it achieves higher Dist-$n$ scores than KGSF and only results in a slight drop compared with RevCore ($+$KG). It demonstrates that reviews compared with KG bring more diversity as a richer and more accessible external source.}

    

\begin{table}[t]
    \centering
    \setlength{\abovecaptionskip}{2mm}
    \setlength{\belowcaptionskip}{2mm}
    \begin{tabular}{lcccc}
    \toprule
    \small \textbf{Models} & \small \textbf{Dist-2} & \small \textbf{Dist-3} & \small \textbf{Dist-4} & \small \textbf{PPL}\\
    \midrule
   \small  Trans  & 0.148  & 0.151 & 0.137 & 17.0\\
    \small Redial & 0.225 & 0.236 & 0.228 & 28.1\\
    \small KBRD   & 0.263 & 0.368 & 0.423 & 17.9\\
   \small  KGSF   & 0.289 & 0.434 & 0.519 & \bf~~9.8 \\
    \hline

    \small RevCore ($-$KG) & 0.373  &  0.527  & \bf  0.615  &   10.7 \\
    \small \textbf{RevCore ($+$KG)} &\bf 0.424 &\bf 0.558 & 0.612 & 10.2\\
    \bottomrule
    \end{tabular}
    \caption{Results on the conversation task. Best results are in bold. }
    \label{tab:conv}
\end{table}

\paragraph{Human Evaluation}
{We adopt human evaluation on a random selection of 100 multi-turn dialogues from the testing set.}
{Given one dialogue context, each generated response is scored ranging from 0 to 1, with a higher value indicating a more coherent, fluent, and informative utterance. The final result is calculated as the average score of three annotators, as summarized in Table~\ref{tab:human}.}
{The proposed RevCore (with or without KG) is consistently better than all the baselines, especially on the metric of informativeness in a large margin.}
{It further proves the effectiveness of our method, and also verifies its superiority in numerical results.}

\begin{table}[t]
    \centering
    \setlength{\tabcolsep}{2mm}
    \setlength{\abovecaptionskip}{2mm}
    \setlength{\belowcaptionskip}{2mm}
    \begin{tabular}{lccc}
    \toprule
    \small \textbf{Models} & \small \textbf{Coherence} & \small  \textbf{Fluency} & \small \textbf{Informat} \\
    \midrule
    \small Trans  & 0.189 & 0.226 & 0.115  \\ 
    \small Redial & 0.225 & 0.455 & 0.228 \\
    \small KBRD   & 0.263 & 0.468 & 0.283 \\
    \small KGSF   & 0.324 & 0.502 & 0.332  \\ 
    \hline
      \small RevCore ($-$KG) & 0.556 & 0.493 &  0.682\\ 
    \small \textbf{RevCore ($+$KG)} &\bf 0.601 &\bf 0.567 & \bf 0.718 \\ 
  
    \bottomrule
    \end{tabular}
    \caption{ Human evaluation results. ``Informat" denotes informativeness. Best results are in bold. }
    \label{tab:human}
\end{table}

\paragraph{Ablation Study}
{We demonstrate the contribution of each part on the conversation task by constructing an ablation study based on three variants of our complete model, including:} 
{(1) RevCore ($-$revCP) by removing the copy mechanism for reviews, (2) RevCore ($-$revRA) by removing the review attention layers from the transformer decoder, and (3) RevCore  ($-$revEN) by removing the sentiment-aware review encoder (the reviews  share the same encoder with the context).}
{As shown in Table~\ref{tab:ablation}, first, all the techniques are useful to improve the final performance in generating diversified utterances.}
{Besides, the copy mechanism and the review attention layers seem to be more important in conversation diversity.} 
{One of the potential reasons is that these two components are directly related to the decoding stage. Separated encoders for review and context lead to a slight increment, which shows that sharing a common encoder is an alternative solution.}

\begin{table}[t]
    \centering
    \setlength{\abovecaptionskip}{2mm}
    \setlength{\tabcolsep}{2mm}
    \begin{tabular}{ccccc}
    \toprule
    \small \textbf{Models} & \small \textbf{Dist-2} &\small  \textbf{Dist-3} &\small  \textbf{Dist-4} & \small \textbf{PPL} \\
    \midrule
    \small \textbf{RevCore ($+$KG)}    &\bf 0.424 &\bf 0.558 & \bf 0.612&   10.2 \\
    \hline
    \small $-$revCP        & 0.353 & 0.443 & 0.503 &  \bf10.0 \\
    \small $-$revRA        & 0.328 & 0.428 & 0.516 & 13.2  \\ 
    \small $-$revEN        & 0.394 & 0.534 & 0.586 & 10.8 \\ 
    \bottomrule
    
    \end{tabular}
    \caption{Ablation study on the conversation task. }
    \label{tab:ablation}
\end{table}

\begin{figure*}[t]
    \centering
    \includegraphics[width=6.3in]{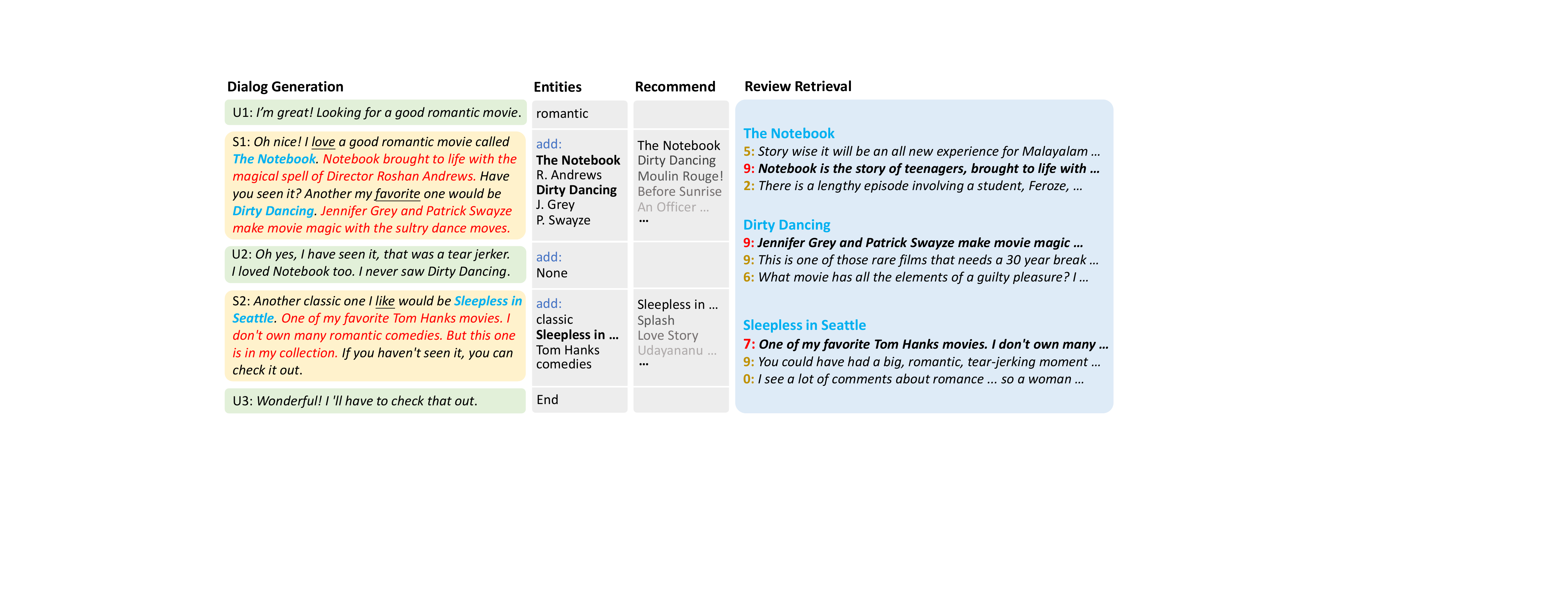}\\
    \caption{ Case study. $U(i)$ (green) and $S(i)$ (yellow) represent user and system, respectively. In the ``Dialogue Generation", items are marked in blue font, explanatory sentences are in red. Items are in bold font in ``Entities" frames. In ``Recommend" frames, a darker color represents a higher probability. ``Review Retrieval" gives retrieved review examples, with their sentiment value (0-9) at the most left, and the selected reviews are in bold. }
    \label{fig:visualize}
\end{figure*}

\subsection{Case Study}
{In this part, we present a visualized example to illustrate how our model works in practice, as shown in Figure~\ref{fig:visualize}.}
{First, the sentiment-aware review retrieval module guarantees the coherence of incorporating reviews to some extent, for example in Figure~\ref{fig:visualize}, negative comments (the last row for the movie \textit{The Notebook}) are filtered out considering the positive attitude in the original utterance.}
{Secondly, incorporated reviews exactly enrich the context for better recommendation.}
{As seen in the first yellow frame, many entities like ``\textit{Roshan Andrews}'' mentioned in the review are added into the entity set.}
{Note that some other entities are also added from the reviews that are incorporated into users' utterances as described in Section~\ref{sec:method}, which is not visualized here but brings recommendation accuracy improvement as well.}
{Last but not least, the generated responses are more informative to use more varied expressions like ``\textit{the magic spell}'' and ``\textit{the sultry dance}''.
Besides, they uncover more details related to the movie that acts as explanatory sentences that make recommendations accepted more easily and naturally.}

\subsection{Discussion}
\label{diss}
\begin{table}[!tp]
  \centering
  \begin{threeparttable}
    \setlength{\tabcolsep}{0.9mm}
    \setlength{\abovecaptionskip}{2mm}
    \begin{tabular}{lcccccc}
    \toprule
    \multirow{2}{*}{\small \textbf{Len}~~} &\multicolumn{3}{c}{\small Recommendation} &\multicolumn{3}{c}{\small Conversation}\cr
    \cmidrule(lr){2-4} \cmidrule(lr){5-7}
         & \small \textbf{R@1} & \small \textbf{R@10} & \small \textbf{R@50} & \small ~~\textbf{Dist-3} & \small ~~\textbf{Dist-4} & \small \textbf{PPL} \\
    \midrule
    10~  & 4.5 & 21.8 & 37.0 &   ~~0.491  & ~~0.564    & \bf~~10.2      \\
    20~  & \bf 6.1 & \bf 23.6 & \bf 45.4  & \bf ~~0.558 & \bf ~~0.612 & \bf~~10.2  \\
    30~  & 4.7 & 22.0 & 41.3 & ~~0.263 & ~~0.426 & ~~13.1 \\
    40~  & 5.8 & 21.3 & 41.3 & ~~0.289 & ~~0.439 & ~~14.0 \\
    50~  & 5.1 & 22.3 & 40.8 & ~~0.304 & ~~0.520 & ~~13.8 \\
    \bottomrule
    \end{tabular}
    \caption{Performance of RevCore when incorporating review sentences with different length (Len). Best results are in bold.}
    \label{tab:length}
    \end{threeparttable}
\end{table}

\paragraph{Longer Review, Better Performance?}  
{In our basic setting, each retrieved review sentence is formed by 20 words.}
{We conduct a series of experiments by setting the length of retrieved review sentences as 10, 20, 30, 40, and 50 to inspect the effect of review length.}
{The results of using different lengths are shown in Table~\ref{tab:length}, among which 20 is the best across all metrics.
An interesting finding is that continually increasing the review length makes no benefits after reaching 20. Over introducing external text may suppress original text, thus 20 is a better choice to keep the balance between them.}

\renewcommand{\arraystretch}{1}
\begin{table}[!tp]
  \centering
  \begin{threeparttable}
    \setlength{\tabcolsep}{0.9mm}
    \setlength{\abovecaptionskip}{2mm}
    \begin{tabular}{lcccccc}
    \toprule
    \multirow{2}{*}{\small Method} &\multicolumn{3}{c}{\small Recommendation} &\multicolumn{3}{c}{\small Conversation}\cr
    \cmidrule(lr){2-4} \cmidrule(lr){5-7}
                     &\small \bf R@1   & \small \bf R@10 & \small \bf R@50  &\bf ~\small Dist-3           & \small ~~\bf Dist-4    & \small \bf PPL\cr
    \midrule
    \small iCorpus          & 2.1   &  8.5  & 20.3  & ~0.430  & ~~0.555  & ~11.2 \cr
    \small R-H-W            & 2.3   & 10.7  & 25.2  & ~0.307  & ~~0.347  & ~10.9 \cr
    \small C-H-W            & 5.4   & 23.3  & 42.9  & ~0.471  & ~~0.559  & ~13.1 \cr
    \small C-H-S            &  4.2  & 22.4  & 39.4  & ~0.534  & ~~0.586  & ~11.4 \cr
    \small C-S-S & \bf 6.1  & \bf 23.6  & \bf45.4  & ~\bf 0.558  & ~~\bf 0.612 & \bf ~10.2 \cr
    \bottomrule
    \end{tabular}
    \caption{Results of various retrieval strategies. Three characters from left to right represent three factors: (i) C/R: correctly/randomly matched movie-review pairs; (ii) S/H: ranking by sentiment value or helpful score respectively; (iii) S/W: sentence/word-wise manner. ``iCorpus" indicates using irrelevant corpus. Best results are in bold.}
    \label{tab:retrival}
    \end{threeparttable}
\end{table}

\paragraph{Appropriate Reviews Help More?}
{The ``review sentence" are obtained from a 3-stage process, namely, searching item-matched reviews from the database, ranking them by the helpful score or sentiment value, and constructing ``review sentence" word-wisely or sentence-wisely.}
{Therefore, we conduct control experiments to inspect these three factors.}
{As shown in Table~\ref{tab:retrival}, (i) using reviews randomly matched with items (R-H-W) results in significantly lower R@$k$ and Dist-$n$ scores; (ii) ranking by sentiment value (C-S-S) leads to better performance across all metrics than by helpful score (C-H-S), which demonstrates the necessity of using sentiment-aware review retrieval; (iii) the sentence-wise manner (C-H-S) gets a lower PPL than the word-wise one (C-H-W), which is reasonable because the incorporated reviews made up of random words causes the fluency loss.}
{Besides, another experiment is conducted to verify the necessity of using a movie-review database. A food-review database is constructed as a topic-irrelevant corpus (iCorpus), which results in the lowest R@$k$ yet not bad Dist-$n$ scores. It shows that despite the response diversity brought by the external corpus, the unrelated entities from another domain have negative impacts on the recommendation accuracy.} 

\section{Related Work}
Recommender systems have emerged as a separate research area and now play an indispensable role in daily social lives.
Traditional recommender systems tend to work statically, primarily relying on content-based approaches or the collaborative filtering hypothesis~\citep{resnick1994grouplens,pazzani2007content,wang2019explainable}, which assumes that similar users may have similar interests.
Afterward, more sophisticated methods using neural networks are proposed and prove effective. For instance, neural factorization machines~\citep{he2017neural} and deep interest networks~\citep{zhou2018deep} are used to estimate user preferences based on historical user-item interactions. Graphs are adopted in \citet{wang2019explainable,wang2019neural} to model complex relations among users, items, and attributes for a better representation of data.

In recent years, major advances made in dialog systems~\citep{dodge2015evaluating,yan-etal-2016-docchat,benni2016human,bordes2016learning,song-etal-2020-summarizing} and structured knowledge-based info-seeking technics including question answering~\citep{bao-etal-2014-knowledge,bao-etal-2016-constraint,yin2015answering,yih-etal-2015-semantic,shao2019weakly} and question generation~\citep{serban-etal-2016-generating,bao2018table,duvsek2020evaluating} have encouraged the development of \textit{conversational recommendation} systems, which dynamically obtain user preferences through interactive conversation with users.
Multiple datasets have been constructed~\citep{dodge2015evaluating,li2018towards,kang2019recommendation} to facilitate the study of this task.
\citet{li2018towards} collect a standard human-to-human multi-turn dialog dataset focusing on providing movie recommendations.
Based on these datasets, various approaches are proposed to address different issues in CR systems.
Specifically, external information is introduced to alleviate the cold-start problem, including knowledge bases~\citep{wang2018ripplenet}, social networks~\citep{daramolaimproving}, and knowledge graphs~\citep{chen2019towards}. \citet{christakopoulou2016towards} use bandit-based explore-exploit strategy to minimize the number of user queries.
\citet{liu2020towards} conduct multi-goal planning to make a proactive conversational recommendation over multi-type dialogues.
A multi-view method is proposed in~\citet{chen2020towards} for the explainable conversational recommendation.
The work of \citet{pecune2020socially} builds a socially aware CR system engaging its users through a
rapport-building dialogue to improve users' perception.

Different from all aforementioned previous work, we offer an alternative to AIG with an augmented conversational recommendation system by incorporating reviews that highly relevant to items.
Particularly, our model is able to learn better user representations from a review-enriched dialogue context, which enables a high-quality recommendation and response generation.

\section{Conclusion}

In this paper, we proposed a novel CR framework with review augmentation, including a sentiment-aware retrieval module, a recommender exploiting the review-enriched user profile,  an encoder for enhancing semantic embedding of selected reviews,  and a review attentive decoder to integrate review information for dialogue response generation.
Experimental results show that our approach achieves consistent and significant improvements of both recommendation and dialogue responding over baselines, and is able to generate informative responses without losing fluency and coherence.

\section*{Acknowledgement}
The work was supported in part by the Key Area R\&D Program of Guangdong Province with grant No. 2018B030338001, by the National Key R$\&$D Program of China with grant No. 2018YFB1800800, by Shenzhen Outstanding Talents Training Fund, by Guangdong Research Project No. 2017ZT07X152, and by the National Key Research and Development Program of China under Grant No. 2018YFB2100802.

\bibliography{acl2020_new}
\bibliographystyle{acl_natbib}

\appendix
\section{Statistics for Conversation Dataset and Reviews\label{appendixA}}
 Conversations in the REDIAL dataset consist of 163,820 utterances, of which 15.80\% have reviews added. The vocabulary size of REDIAL is increased by 13.14\% (from 23,356 to 26,427). Among the mentioned 6,927 movies in all conversations, 40\% of them are randomly chosen and linked with reviews to keep the balance between the original source and external source. We count the ratio of ``disliked" movies by the recommender to explain the improvements brought by doing sentiment-aware retrieval when incorporating reviews. We also show the ratio of unseen movies by the recommender to show the need of introducing reviews to ``talk more".  Comprehensive statistics are listed in Table~\ref{tab:sta}. 
 \begin{table}[t]
    \centering
    \setlength{\abovecaptionskip}{2mm}
    \setlength{\tabcolsep}{1.5mm}
    \begin{tabular}{lr|lr}
    \toprule
    \textbf{Items} & \textbf{Statistic} & \textbf{Items} & \textbf{Statistic} \\
    \midrule
    Dialogues              &  10,021 & Cnd Movies             &  64,368  \\
    Utterances             & 163,820 & Mnt Movies             &   6,924  \\ 
    $+$reviews &  25,884 &  $+$reviews &   2,711  \\ 
    \midrule
    Disliked               &  ~4.9\% & Not Seen               &  31.9\%  \\
    Liked                  &  81.2\% & Seen                   &  61.3\% \\
    Did not say            &  13.9\% & Did not say            &  ~6.8\% \\
    \bottomrule
    
    \end{tabular}
    \caption{Statistics for the REDIAL dataset with incorporated reviews. ``Cnd" denotes ``Candidate", and ``Mnt" denotes  ``Mentioned" to indicate the movies that mentioned in the conversations. }
    \label{tab:sta}
\end{table}

\section{Experiment Details\label{appendixB}}

\paragraph{Hyper-parameter Settings} For a fair comparison, most hyper-parameters are kept consistent with KGSF.
We did not search for more hyper-parameters combinations to achieve additional improvements apart from our main idea. The shared hyper-parameters include: embedding dimension set as 128 in the recommender component and 300 in the dialogue component, the layer number of both GNN in the KG module as 1, the batch size as 32, word embedding initialization via word2vec\footnote{\url{https://radimrehurek.com/gensim/models/word2vec.html}}, the optimizer as Adam, the learning rate as 0.001, the epoch number as 30, etc.

\begin{table}[t]
    \centering
    \setlength{\abovecaptionskip}{1mm}
    \setlength{\tabcolsep}{1mm}
    \begin{tabular}{lccc}
    \toprule
    \textbf{Models} & \textbf{Param} & \textbf{Tra Time} & \textbf{Inf Time} \\
    \midrule
    KGSF~~                 &  130.51    & ~~4979.88  &  65.30 \\
    RevCore ($-$KG)~~      &  ~~71.12   & ~~1045.65  &  59.57  \\
    RevCore ($+$KG)~~      &  133.30    & ~~3308.28  &  51.46  \\
   
    \bottomrule
    
    \end{tabular}
    \caption{Comparison of three models on the number of parameters (million), training (``Tra") time for 30 epochs (second), and inference (``Inf") time (second).}
    \label{tab:size}
\end{table}

\paragraph{Training Strategies} To train the whole model, three steps are included: (i) pre-training the sentiment predictor in the review retrieval module; (ii) training the recommender component by minimizing $\mathcal{L}_{rec}$; (iii) training the dialogue component by minimizing $\mathcal{L}_{gen}$. In the first step, the predictor takes each sentence in the review as input and outputs the sentiment, with the corresponding rating set as the label. In the second and third steps, our implementation refers to the training algorithm for the KGSF model. It first pre-trains the parameters in KG for entity representation by minimizing the Mutual Information Maximization loss between two KG embedding, then trains the recommender component by minimizing the recommendation loss and also updating the parameters in the KG module, and finally the dialogue component by minimizing the generation loss with all other modules' parameters ``frozen".

\section{Model Size and Running Speed\label{appendixC}}
The model size and running speed of KGSF, RevCore ($+$KG), and RevCore ($-$KG) are all listed in Table~\ref{tab:size}.
Note that all three models are implemented with Pytorch\footnote{\url{https://pytorch.org/}}, trained for 30 epochs, and experimented on NVIDIA A100-SXM4 for 5 times to compute the average running time.

\section{Results on the Validation Set\label{appendixD}}
We present the validation result of RevCore with and without KG on the REDIAL dataset as a reference for reproducing. All validation results are shown in Table~\ref{tab:val}, with test results as well.

\begin{table}[t]
  \centering
  \begin{threeparttable}
    \setlength{\tabcolsep}{3mm}
    \setlength{\abovecaptionskip}{2mm}
    \scalebox{0.93}{
    \begin{tabular}{l|cc|cc}
    \toprule
    \multirow{2}{*}{Metrics} &\multicolumn{2}{c|}{RevCore ($+$KG)} &\multicolumn{2}{c}{RevCore ($-$KG)}\cr
    \cmidrule(lr){2-3} \cmidrule(lr){4-5}
                         & \bf Val  & \bf Test   & \bf Val  & \bf Test  \cr
    \midrule
    R@1           & ~~6.13  & ~~6.11 & ~~4.55  & ~~4.19 \cr
    R@10          & 23.49   & 23.62  & 23.35   & 22.71  \cr
    R@50          & 40.65   & 45.43  & 45.33   & 43.28  \cr
    \midrule
    Dist-2        & 0.418   & 0.424  & 0.410   & 0.373  \cr
    Dist-3        & 0.582   & 0.558  & 0.582   & 0.527  \cr
    Dist-4        & 0.675   & 0.612  & 0.668   & 0.615  \cr
    PPL           & 10.89   & 10.24  & 10.14  &  10.69    \cr
    \bottomrule
    \end{tabular}
    }
    \caption{Validation (Val) and test results on the REDIAL dataset of RevCore ($+$KG) and RevCore ($-$KG).}
    \label{tab:val}
    \end{threeparttable}
\end{table}

\begin{table}[t]
  \centering
  \begin{threeparttable}
    \setlength{\tabcolsep}{2.1mm}
    \setlength{\abovecaptionskip}{2mm}
    \scalebox{0.98}{
    \begin{tabular}{l|cc|cc}
    \toprule
    \multirow{2}{*}{Metrics} &\multicolumn{2}{c|}{RevCore ($+$KG)} &\multicolumn{2}{c}{RevCore ($-$KG)}\cr
    \cmidrule(lr){2-3} \cmidrule(lr){4-5}
                         & \bf Mean  & \bf Devi   & \bf Mean  & \bf Devi  \cr
    \midrule
    R@1           & ~~5.70  & ~$\pm$ 0.67 & ~~3.75  & ~$\pm$ 0.13 \cr
    R@10          & 22.80   & ~$\pm$ 1.81  & 21.53   & ~$\pm$ 0.68  \cr
    R@50          & 40.75   & ~$\pm$ 2.18  & 44.68   & ~$\pm$ 0.43  \cr
    \midrule
    Dist-2        & 0.394   & $\pm$ 0.039  & 0.373   & $\pm$ 0.031  \cr
    Dist-3        & 0.551   & $\pm$ 0.062  & 0.527   & $\pm$ 0.051  \cr
    Dist-4        & 0.633   & $\pm$ 0.073  & 0.616   & $\pm$ 0.062  \cr
    \bottomrule
    \end{tabular}
    }
    \caption{Mean and deviation of recall rates (\%) and distance scores for RevCore ($+$KG) and RevCore ($-$KG).}
    \label{tab:mean_dev}
    \end{threeparttable}
\end{table}

\section{Mean and Standard Deviation\label{appendixE}}
We implement the major experiment 4 times to inspect the mean and standard deviation of the performance of RevCore across all metrics. The reported results in the paper of both recommendation accuracy and conversation quality are the mean results. Results are shown in Table~\ref{tab:mean_dev}.

\end{document}